\documentclass[11pt]{article}

\usepackage[preprint]{acl}

\usepackage{times}
\usepackage{latexsym}
\usepackage{amssymb}
\usepackage{amsmath}
\usepackage[capitalize]{cleveref}

\usepackage{threeparttable}
\usepackage[T1]{fontenc}

\usepackage[utf8]{inputenc}

\usepackage{microtype}

\usepackage{inconsolata}

\usepackage{graphicx}
\usepackage[normalem]{ulem}

\usepackage{booktabs}   
\usepackage{tabularx}   
\usepackage{ulem}       
\usepackage{subcaption}
\definecolor{mygreen}{RGB}{4, 135, 76}
\definecolor{myblue}{RGB}{35, 80, 153}

\newcommand{\draftcomment}[3]{{\textcolor{#3}{[#1]#2}}}

\newcommand{\roy}[1]{\draftcomment{#1}{\textsc{roy}}{purple}}
\newcommand{\rs}[1]{\roy{\sout{#1}}}
\newcommand{\ra}[1]{\textcolor{purple}{[#1]}}
\newcommand{\rr}[2]{\rs{#1}\ra{#2}}

\renewcommand{\ra}[1]{#1}
\renewcommand{\rs}[1]{}

\usepackage{xspace}
\newcommand{\mknn}[0]{m-KNN\xspace}
\newcommand{\ssm}[0]{SSM\xspace}
\newcommand{\ssms}[0]{SSMs\xspace}

\title{Global Divergence, Local Convergence:\\ Representation Geometry in SSMs and Transformers}

\author{
  \textbf{Amit Ben-Artzy} \quad \textbf{Roy Schwartz} \\
    Computer Science, The Hebrew University of Jerusalem \\
  \texttt{\{amit.benartzy, roy.schwartz1\}@mail.huji.ac.il}
}

\begin{document}
\maketitle
\begin{abstract}

Recent state-space models~(\ssms) such as Mamba achieve language modeling performance comparable to transformers despite relying on fundamentally different architectures. This raises an important question: how do these structural differences influence the geometry and functional nature of their internal representations? We study this question through a multi-scale analysis of representations in transformers, \ssms, and hybrid architecture. First, we find that SSMs distribute their representational variance much more uniformly across dimensions compared to transformers, which are heavily dominated by a single principal direction~(\cref{fig: pca_plot}). By evaluating a hybrid architecture, we observe that the representation space becomes increasingly skewed toward a single dominant direction after each attention layer.
Next, we explore how the different geometric spread of representations impacts representational capacity through compressibility. Surprisingly, we find that despite their contrasting geometric structures, both architectures exhibit tightly matched effective capacities. We further investigate whether this skewed geometry affects how concepts are encoded. Using rank-constrained probes, we demonstrate that both architectures encode concepts in subspaces of surprisingly similar dimensionality. Moreover, we demonstrate that the transformers' dominant principal direction does not inherently encode more conceptual information. Finally, we zoom in and examine the alignment between manifolds, either by analyzing representations of specific topics or by looking at the nearest neighborhoods of tokens, and find that they are highly aligned. Ultimately, our analysis suggests that while transformers and \ssms induce different usage of latent space, they display a striking structural alignment at the level of local semantic manifolds.\footnote{Code is available at \url{https://github.com/schwartz-lab-NLP/ssm-vs-transformer-geometry}}

\end{abstract}

\begin{figure}[ht!]
    \centering
    \includegraphics[width=1\linewidth]{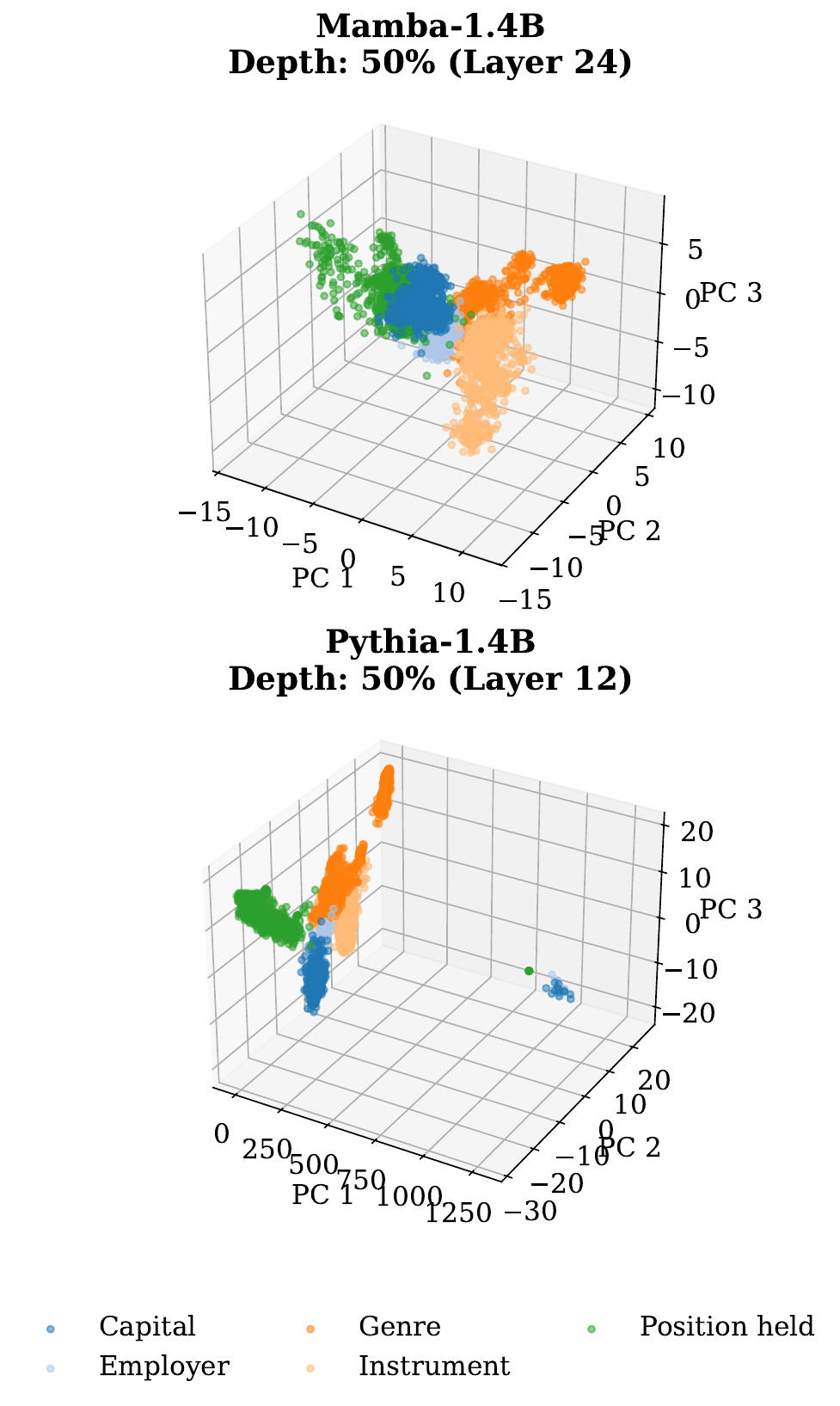}
    \caption{PCA of representations from Pythia and Mamba over a subset of the LAMA TREx  dataset. Mamba's representations are more spread out across the space.} 
    \label{fig: pca_plot}
\end{figure}

\section{Introduction}

State-space models (SSMs) such as Mamba~\cite{mamba1} have emerged as highly efficient alternatives to transformers~\cite{attention_all_you_need}, achieving competitive empirical performance on a variety of language modeling tasks \cite{empirical_mamba_study}. While transformers rely on global attention mechanisms, SSMs compress context into a bounded state, resulting in very different architectural priors.\footnote{We use \ssms to refer specifically to the Mamba architecture, avoiding ambiguity with the Mamba model family.}

This divergence in computational paradigms raises a  key question: do these distinct architectures induce fundamentally different internal representations? To investigate this systematically, we conduct a multi-scale analysis of representations across both model families. To ensure a direct and structurally equivalent comparison, we focus our analysis entirely on the residual stream, analyzing \ra{several }representative models: two transformer models~(Pythia,~\citealp{pythia}; and Falcon,~\citealp{falcon}); two \ssm models~(Mamba ~\citealp{mamba1} and FalconMamba,~\citealp{falcon_mamba}) and one hybrid model, alternating Mamba and transformer layers~(Jamba2,~\citealp{jamba}).\footnote{Results for more models are provided in the appendix.}

We first examine the global geometry of the models over a general corpus~(\cref{sec:global_geom}). Specifically, we measure effective dimensionality via RankMe~\cite{rank_me} and the degree of isotropy via IsoScore~\cite{rudman-etal-2022-isoscore}---the extent to which token embeddings are uniformly distributed across all directions in space. We find a stark contrast: \rr{Mamba's}{\ssms'} residual stream is highly isotropic and exhibits a high effective dimensionality. In contrast, transformers are notoriously anisotropic~\cite{Gao2019Jul,Godey2024Jan}, characterized by a dominant first principal component (PC1) that explains a disproportionately large share of the variance. We further validate this trend by examining the hybrid model Jamba2, where we find that after each attention layer, there is a sharp shift toward anisotropy.

Given these contrasting global geometries, we investigate whether the two architectures use their representation spaces differently to encode specific concepts (\cref{sec:intrinstic_dim}). Using autoencoders and rank-constrained probing \cite{low_dim_probe} to measure the intrinsic dimensionality required for decoding, we find that they exhibit similar functional dimension. 
Furthermore, we find that despite transformers' high alignment with a single dominant direction---it does not capture more information than \rr{in Mamba's}{\ssms'} isotropic representations. To understand whether concept information is concentrated within specific geometric dimensions, we systematically ablate the top-$k$ principal components across both architectures. We found that both models experience a comparable decline in probe accuracy, demonstrating that these concepts are not exclusively localized in the dominant dimensions.
Next, we evaluate structural similarity in  \cref{sec:local_similarity}. First, we apply whitened centered kernel alignment~(CKA, \citealp{cka_1,cka_2}) and find that despite diverging global geometries, the local manifolds of specific concepts exhibit high representational similarity.  Zooming into the token level, we evaluate the structural equivalence of local neighborhoods using mutual $k$-nearest neighbors~(\mknn, \citealp{Wolfram2025Apr}) and find high alignment across architectures.

Finally, we investigate the stabilization dynamics of these properties during training in \cref{sec:temportal_alignement}. An analysis of the transformer checkpoints demonstrates that local properties, such as CKA and \mknn similarity, converge rapidly—within the first 4\% of training. In contrast, global geometry stabilizes significantly later.

Overall, our contributions are as follows:
\begin{itemize}
    \item \textbf{Global Geometric Divergence:} We provide a systematic characterization of the representation geometry in \ssms.  We demonstrate that the residual streams of transformers and SSMs exhibit substantially different global geometries~(anisotropic vs.~isotropic).

    \item \textbf{Local convergence:} We reveal that locally, representations are highly aligned: both architectures rely on similar intrinsic dimensionalities for concept decoding and exhibit highly correlated local token neighborhoods and concept manifolds.
    \item \textbf{Dynamic mechanisms \& hybrid architectures:} By tracking intermediate checkpoints, we show that local structural convergence emerges early in pretraining. Furthermore, we reveal that hybrid architectures (Jamba2) exhibit representations that  oscillate between the isotropic and anisotropic geometries of their constituent layers.
\end{itemize}

\begin{figure*}[ht!]
    \centering

    \includegraphics[width=1\textwidth]{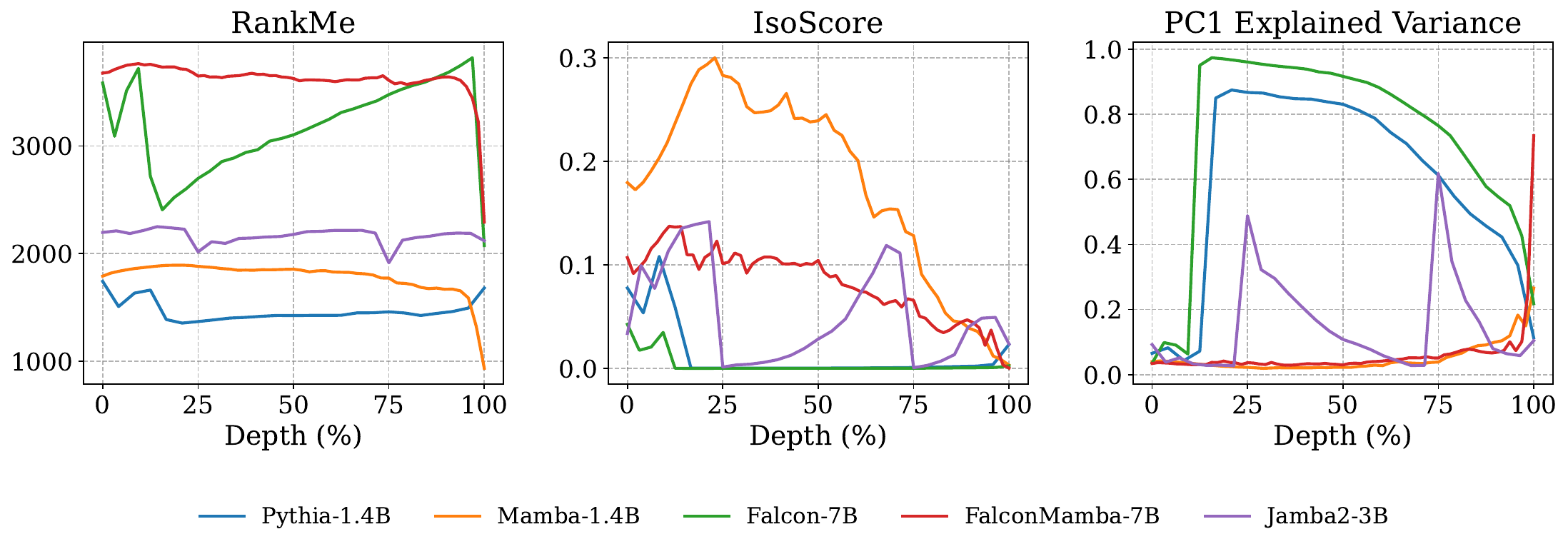}
    \caption{Layer-wise geometric analysis of transformer, State-Space and hybrid models. Metrics calculated using $\approx$500k representations from Fineweb-Edu. \ssm models~(Mamba and FalconMamba) maintain higher effective dimensionality~(RankMe) and more uniform variance distribution~(IsoScore) than the transformer baselines~(Pythia  and Falcon). In Jamba2, attention blocks consistently increase the dominance of the first principal component (PC1) which steadily declines across subsequent Mamba layers.}
    \label{fig:rankme_trained_vs_random}
\end{figure*}

\section{Background: Transformer and SSM Architectures}
\label{sec:arch}
In this work we study the geometrical differences between transformers and \ssms. We begin by describing both architectures. We view each architectural layer as a direct update to the residual stream.

\textbf{Transformer}~\citep{attention_all_you_need} block consists of interleaved attention and feed-forward (FFN) blocks. Given a hidden state $x_l$ at layer $l$, the residual stream is updated as:
\begin{equation}
    x_{l+1} = x_l + \text{Transformer}(x_l)
\end{equation}

The \textbf{Mamba} architecture \citep{mamba1} replaces the attention mechanism with a Selective state space model (SSM). The residual stream is updated as follows:
\begin{equation}
    x_{l+1} = x_l + \text{Mamba}(x_l)
\end{equation}
Internally, a Mamba block consists of an input projection, a 1D depthwise convolution, and a selective SSM S6 mixer, followed by an output projection. The SSM maintains a recurrent hidden state that evolves along the sequence dimension rather than across layers. 

Importantly, our analysis operates at the level of the layer-wise residual stream ${x_l}$, and does not investigate the internal sequence-level state maintained within the Mamba block. This allows us to compare Transformers and SSM-based models within a shared representation framework defined by residual stream dynamics.

\subsection{Models}\label{sec:models}
In our experiments, we evaluate models  from two transformer model families: Pythia~\cite{pythia} and Falcon~\cite{falcon};\footnote{Results for Llama3~\cite{llama_3}, and Qwen2~\cite{qwen_2} are shown in the appendix.} and two SSM model families: Mamba \cite{mamba1} and FalconMamba~\cite{falcon_mamba}. The shared tokenization scheme between these suites facilitates controlled, one-to-one comparisons. Our analysis also evaluates Jamba2~\cite{jamba}, a hybrid model that interleaves Mamba blocks and Transformer blocks.

\section{Global Geometry of Representations}\label{sec:global_geom}

Our goal in this work is to study the different geometrical properties of transformers and \ssms. We start by analyzing their effective dimensionality across layers. Particularly, we ask how many hidden state dimensions are being actively used by each architecture.
To do so, we consider a matrix $X \in \mathbb{R}^{N \times d}$ containing the hidden representation (of dimension $d$) of $N=$ tokens from a given layer $\ell$, and compute three complementary metrics. First, we compute \textbf{RankMe}~\cite{rank_me}, which measures the effective rank as the Shannon entropy of the normalized singular value distribution $p_i = \sigma_i / \sum_j \sigma_j$ derived from the singular values $\sigma_i$ of the representation matrix $X$. To assess global structure, we employ \textbf{IsoScore}~\cite{rudman-etal-2022-isoscore}, which quantifies isotropy by measuring the distance between the data covariance and the identity matrix; a score of 1 indicates that representations use all available dimensions equally. Finally, we report the \textbf{PC1 explained variance}. A high proportion of variance captured by the first principal component suggests representation collapse into a dominant direction, signaling significant anisotropy. We compute each metric on Pythia-1.4B, Falcon-7B, Mamba-1.4B,  FalconMamba-7B, and Jamba2-3B.\footnote{Results for other model families and sizes are presented in~\cref{fig:appendix_metrics}, and show a similar trend.}

We compute hidden representations from all layers of all models, using  $\approx$500K representations obtained over FineWeb-Edu~\cite{fineweb_data}, a curated pre-training corpus filtered for high-quality educational content from the broader FineWeb dataset.

As illustrated in \cref{fig:rankme_trained_vs_random}, SSM-based models  sustain a higher RankMe than their transformer counterparts of the same size across all but the final layers. Additionally, transformer models exhibit substantially lower IsoScores and higher PC1 explained variance ~(regardless of model size). Together, these metrics indicate that \ssm models learn a highly isotropic, uniformly distributed latent representation, whereas transformer representations are highly anisotropic, collapsing into a cone-like geometry dominated by a few principal directions.

To determine whether this anisotropy is specifically tied to the attention mechanism, we analyze the hybrid model Jamba2. In Jamba2, attention layers consistently induce a drop in RankMe alongside an increase in both IsoScore and PC1 explained variance, driving the representations toward a Transformer-like geometry. Conversely, subsequent Mamba layers steadily reverse this trend, restoring the representational structure to one more characteristic of \ssms. This strongly suggests the attention mechanism as the primary driver of dimensional collapse within the network, consistent with prior observations \cite{Godey2024Jan}.

\section{Quantifying the Effective Dimensionality of Representations} \label{sec:intrinstic_dim}

In \cref{sec:global_geom}, we investigated the spatial distribution of representations across $\mathbb{R}^d$, and observed that SSM models feature a distributed latent space while transformers display high anisotropy. Next, we explore how the different geometric spread impacts the capacity of representations. Following this, we assess whether the structural skewness of transformers along a dominant axis correlates with increased information capacity.

\subsection{Model Capacity Limits}\label{subsec:autoencoder}
While isotropy provides insight into the geometric structure of representations, it does not capture how much information the representations encode. To move beyond geometry, we evaluate representational capacity through compressibility. We train a two-layer Autoencoder~(AE) for each layer across various rank constraints $r$. These models are optimized using an $L_2$-reconstruction loss. For evaluation, we measure the Kullback-Leibler~(KL) divergence between the model's original output distribution and the distribution obtained when replacing a single native layer representation of a single token $x_l$  with the AE reconstructions $\text{AE}(x_l)$. The AEs are trained on roughly 4 million tokens from the FineWeb-Edu corpus. 

As illustrated in \cref{fig:constrain_probe_LAMA}(a), restricting the bottleneck rank $r$ induces a highly similar degradation in downstream KL divergence for both architectures across all layers, other than the last SSM layer which is much more sensitive. This similarity  demonstrates that their effective representational capacity is tightly matched, despite the varying geometrical measures and varying architectures.

Additional results for more rank constraint values are shown in \cref{fig:ae_kl_full}, \cref{fig:ae_kl_seed} of \cref{subsec: appendix_ae_full}.

\subsection{The Subspace Geometry of Concepts}

\label{subsec:ablated_probing}
\begin{figure*}[ht!]
    \centering
    \includegraphics[width=1\linewidth]{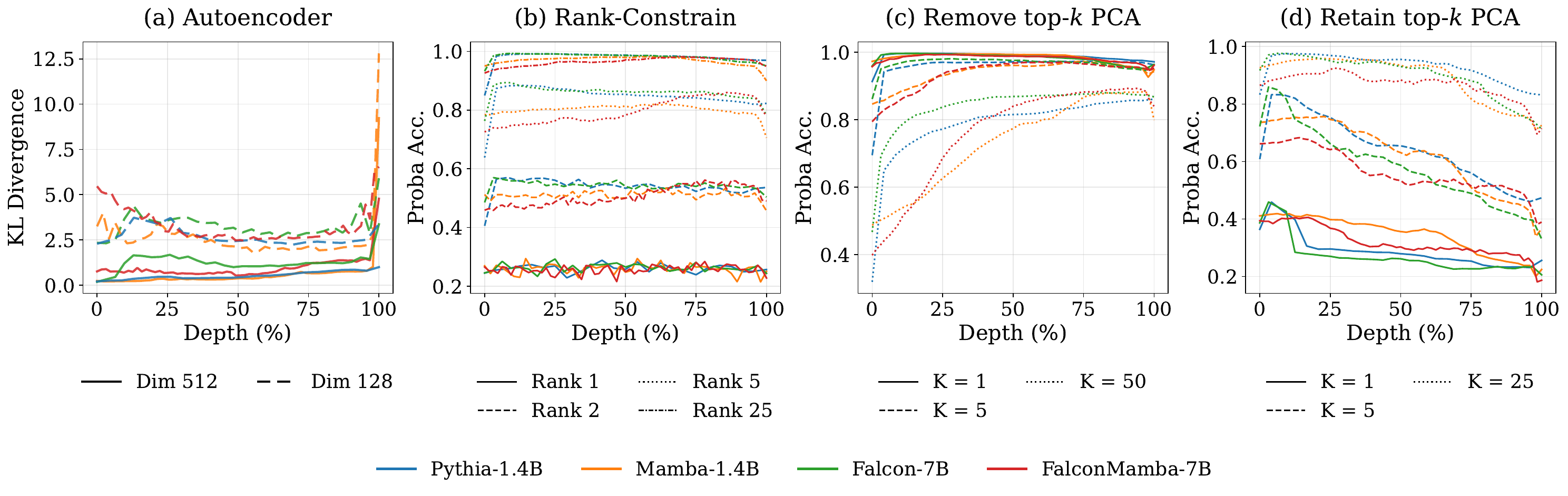}
    \caption{Probing under rank and PC ablations. Left to Right: (a) Original layer representations $x_l$ are replaced by their reconstructions $\text{AE}(x_l)$ using a two-layer autoencoder with varying bottleneck sizes. We then compute the KL divergence between the original and reconstructed distributions to measure information loss. Both architectures exhibit highly similar performance degradation across ranks, indicating tightly matched effective representational capacity, except for the more sensitive final SSM layer. (b) Average rank-constrained probe accuracy across LAMA-TREx relations. Similar rank constraints yield comparable decoding accuracies across model types, despite significant differences in representation geometry. (c) Probe accuracy after removing or (d) retaining the top-$k$ PCA components; results remain consistent across models despite transformers' higher anisotropy.}

    \label{fig:constrain_probe_LAMA}
    
\end{figure*}

We have so far observed that on the one hand SSM representations are uniformly distributed, while transformers are highly anisotropic; and on the other that both architecture behave remarkably similarly when constraining their rank. We now turn to ask how localized individual concepts are within the representation space.

To answer this, we shift our focus from the global token distribution to specific, factual concept manifolds. We evaluate the models using the TREx split of the LAMA dataset \cite{lama_dataset}, which consists of Wikidata triples—comprised of a subject, relation, and object—aligned with corresponding natural language sentences from Wikipedia. Specifically, we present the models with these texts and extract the hidden states from each layer at the final token of the object. Samples from this dataset are available in \cref{sec: lama_trex_samples}.

This methodology allows us to isolate and analyze the latent manifolds of diverse semantic relations, such as occupation, genre, and capital. We leverage this constrained setting to perform rank- and PCA-constrained probing, allowing us to characterize the intrinsic dimensionality required to encode localized relational knowledge across architectures.

\paragraph{Intrinsic dimensionality of relational facts}
How does the global distribution of representations across $\mathbb{R}^d$ influence the spatial localization of distinct concepts and the minimal subspace dimensionality needed for their extraction?

To measure the dimensional footprint of localized concepts, we decode relational facts using rank-constrained probes.
We factorize the probe $W \in \mathbb{R}^{d \times m}$ as $W = AB$, where $A \in \mathbb{R}^{d \times r}$ and $B \in \mathbb{R}^{r \times m}$ are down-projection and up-projection matrices, respectively, with $r$ representing the rank constraint and $m$ denoting number of classes for the object in the LAMA TREx dataset. As shown in~\cref{fig:constrain_probe_LAMA}(b), applying identical rank constraints yields closely matched decoding accuracies across both architectures and all layers. This indicates that these relational facts are embedded within subspaces of similar rank requirements.

\paragraph{Robustness to spectral ablation}
Do the transformer dominant variance directions actually encode more semantic information than SSMs' distributed components? We test this using PCA-constrained decoding.
For each LAMA TREx relation, we compute the principal components of the representation space and evaluate the degradation of probing performance after either retaining or removing the top-$k$ PCA directions. 
First, we see that despite the transformer models' massively dominant PC1 variance, removing the top PCA directions leads to a generally similar performance degradation across architectures~(\cref{fig:constrain_probe_LAMA}(c)).
Second, we see that retaining the top-k PCA dimension degrades the probe at similar rates, across both architectures and scales~(\cref{fig:constrain_probe_LAMA}(d)). This suggests that the dominant directions in the global geometry of transformers do not serve as the primary axes for factual knowledge, but instead reflect broader structural properties of the transformer representation space.

\begin{figure*}[ht!]
    \centering
    \includegraphics[width=1\textwidth]{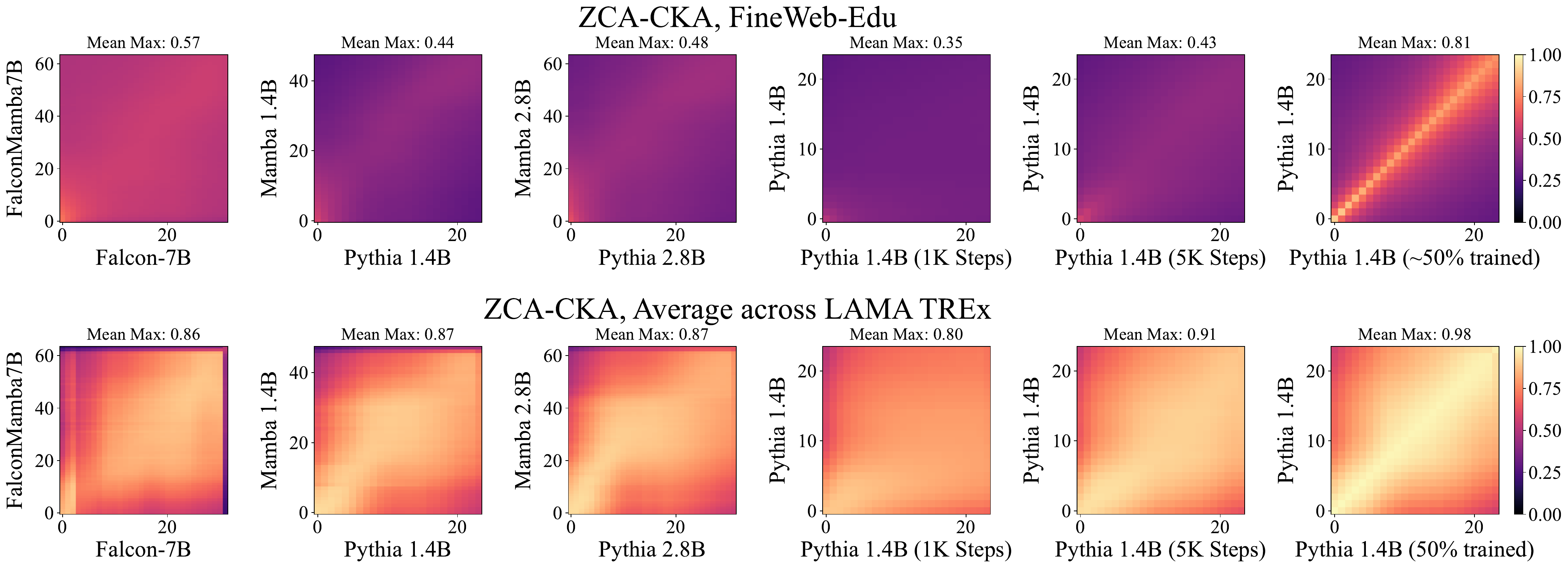}
    \caption{Layer-wise ZCA-CKA alignment on (top) FineWeb-Edu and (bottom) averaged WikiData relations from LAMA TREx. ``Mean max'' indicates the mean of the maximum CKA alignment across layers. Models show pronounced differences over the general corpus but exhibit high structural similarity when restricted to specific relations.}

    \label{fig:cka_combined}
\end{figure*}

\section{Alignment of Semantic Concepts and Local Neighborhoods}\label{sec:local_similarity}

Building upon the global geometric differences identified in \cref{sec:global_geom} and the equivalent intrinsic ranks observed in \cref{sec:intrinstic_dim}, we turn our attention to the direct alignment of their representational spaces, quantifying the extent to which these diverging architectures preserve structural equivalence.
\subsection{Alignment of Concept Manifolds}\label{subsec:cka}

Following the observation that transformers and SSMs compress factual knowledge into subspaces of equivalent dimensionality, we investigate whether these subspaces are also structurally aligned. We quantify this alignment using centered kernel alignment~(CKA;~\citealp{cka_1, cka_2}), which evaluates representational similarity by comparing the inner product structures of different feature spaces.

Linear CKA is heavily weighted by the leading eigenvalues of the covariance matrix, making it highly sensitive to the underlying variance profile of the representations.  This introduces an artifact when comparing transformers---which retain the high anisotropy---against \ssms, which are highly isotropic. We mitigate this discrepancy by applying zero-phase component analysis (ZCA), standardizing both spaces to enable a direct structural comparison.

ZCA transforms representations by applying the inverse square root of their covariance matrix ($\Sigma^{-1/2}$). This transforms representations such that their global covariance becomes the identity matrix, thereby reducing the influence of anisotropic second-order statistics on similarity estimates.
Crucially, for each model and layer, we compute $\Sigma$ globally using 10K matched token representations from FineWeb-Edu. Relation-wise whitening would normalize away relation-specific covariance structure, making all local manifolds isotropic by construction and thereby trivializing geometric comparisons.

For relation-specific LAMA mean-centered representations $X$ and $Y$ from the two models, the globally ZCA-whitened representations are:
$$ \tilde{X} = X \Sigma_{X}^{-1/2}, \quad \tilde{Y} = Y \Sigma_{Y}^{-1/2} $$
We then compute linear CKA on these whitened local manifolds using the Frobenius norm formulation:
$$ \text{CKA}(\tilde{X} \tilde{X}^\top, \tilde{Y} \tilde{Y}^\top) = \frac{\|\tilde{X}^\top \tilde{Y}\|_F^2}{\|\tilde{X}^\top \tilde{X}\|_F \|\tilde{Y}^\top \tilde{Y}\|_F} $$
This global whitening allows CKA to better reflect relation-specific geometric similarity.

\paragraph{Manifold alignment results}

Analysis via ZCA-CKA reveals a clear distinction between global and local manifold geometries (\cref{fig:cka_combined}). Specifically, representational similarity between models evaluated on the same relation is substantially higher than the baseline similarity observed across the global corpus. Ultimately, despite exhibiting divergent global geometries, both architectures arrive at structurally analogous sub-manifolds when compressing specific factual concepts.

\subsection{Neighborhoods Level Alignment}\label{subsec:mknn}

\begin{figure}[h!]
    \centering
    \includegraphics[width=\linewidth]{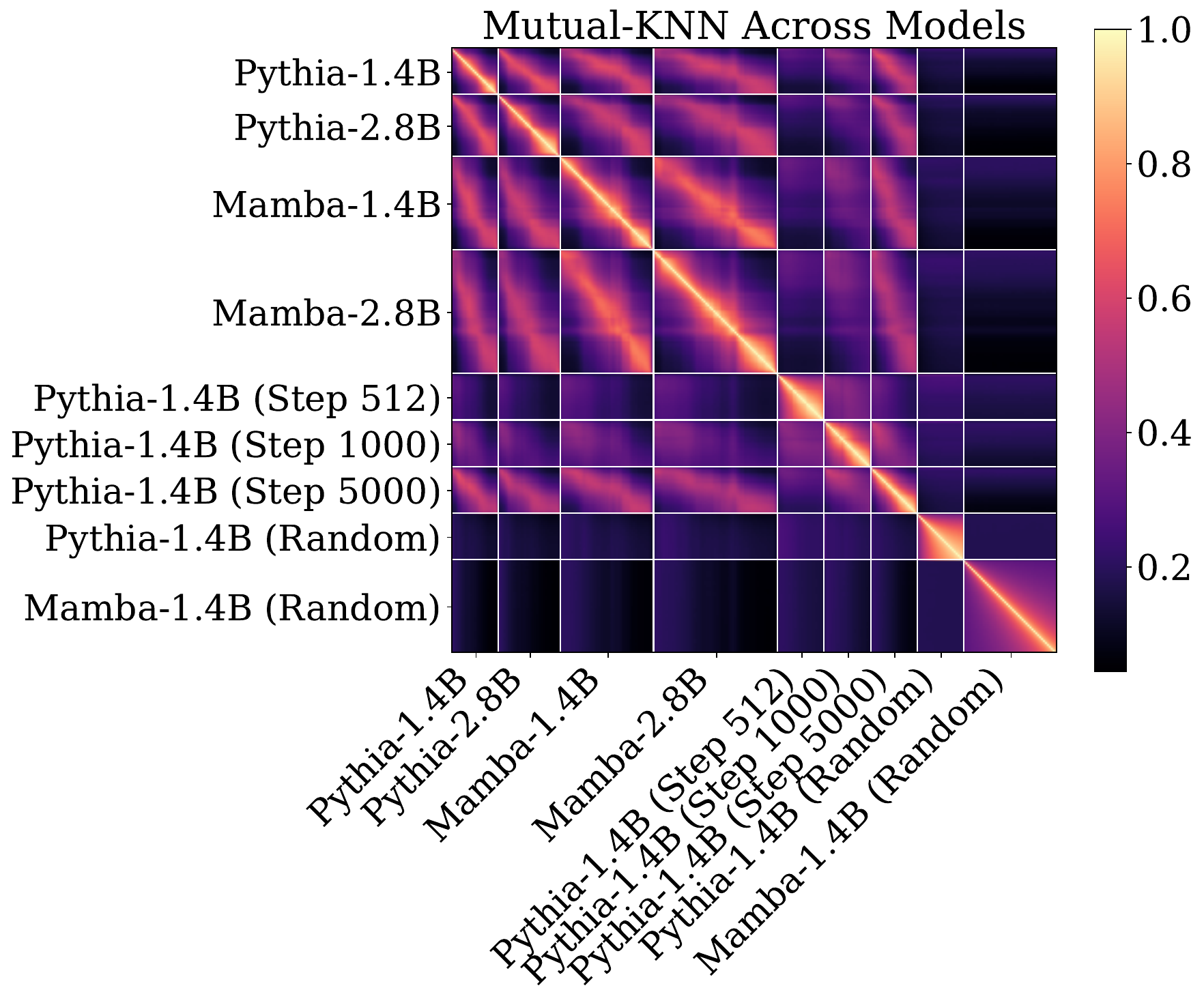}
    \caption{\mknn across models and layers, including random initializations and Pythia's checkpoints. We find that the local manifold of representations is similar across models in similar depths; and this similarity is achieved at Pythia's checkpoint 5K.}
    \label{fig:mknn}
\end{figure}

Having established that topic-specific manifolds align despite global geometric differences, we now examine the representation space at its absolute finest granularity: the local neighborhoods of individual tokens. Specifically, we ask: do representations from transformers and SSMs cluster semantic information similarly at the token level?

To this end, we use the mutual-KNN metric. For token representations $x_i \in X$ and $y_i \in Y$, with $k$ nearest neighborhoods $\mathcal{N}_k^X(x_i)$ and $\mathcal{N}_k^Y(y_i)$, the overlap across $N$ tokens is:
\begin{equation}
    \text{\mknn@}k = \frac{1}{N} \sum_{i=1}^{N} \frac{|\mathcal{N}_k^X(x_i) \cap \mathcal{N}_k^Y(y_i)|}{k}
\end{equation}

As all models we consider share an identical tokenizer, we can directly compute the nearest-neighbor overlap for identical tokens across their respective representation spaces. We define nearest neighbors with cosine similarity.
We measure mutual k-Nearest Neighbors (\mknn) on 10,000 token samples from FineWeb-Edu  and $k=10$. Importantly, \mknn relies entirely on relative distances; it is invariant to the global linear transformations and scaling disparities.

\paragraph{Transformers and \ssms converge}
As illustrated in \cref{fig:mknn}, the \mknn  overlap between fully trained transformer and \ssm models is remarkably high across corresponding depths. This demonstrates that despite architectural differences, both models cluster individual tokens into strongly aligned microscopic neighborhoods. Furthermore, while these local topologies evolve significantly as representations propagate forward through the residual stream, transformers and \ssms track this evolution in parallel, maintaining high mutual overlap at equivalent network depths. Additional results for Falcon-7B and FalconMamba-7B are shown in \cref{fig:mknn_falcon} (\cref{sec:mknn_falcon}).

\section{Temporal Dynamics of Alignment}\label{sec:temportal_alignement}

We next investigate \textit{when} these representational similarities emerge by analyzing Pythia's intermediate checkpoints.\footnote{Checkpoints for the other models are unavailable, so we anchor our temporal analysis on Pythia's trajectory.}

We observe that the local structures align remarkably early. As shown by CKA (\cref{fig:cka_combined}) and \mknn (\cref{fig:mknn}), Pythia's representations match Mamba's after just 5,000 training steps ($\sim$4\% of training). At this point, cross-architecture and cross-scale similarity are already high, matching Pythia's self-similarity to its own final state. This rapid saturation occurs in both general and concept-specific contexts, though absolute similarity is higher in the latter.

In contrast, global geometry matures much slower (\cref{fig:pythia_checkpoints}). Global metrics like IsoScore and PC1 Explained Variance do not stabilize until approximately 25\% of training. Furthermore, while these metrics show monotonic convergence, the intrinsic dimensionality (RankMe) fluctuates non-monotonically—first increasing, then decreasing. This suggests that while local neighborhoods lock in almost immediately, the global space undergoes structural expansion and compression before settling.

\section{Related Work}

\paragraph{Representational convergence}
The inquiry into whether distinct neural networks arrive at shared internal representations originated with foundational work by \citet{Li2015Nov}, who demonstrated that while independently trained models converge onto identical semantic subspaces, they distribute this information across entirely different individual basis vectors. Extending this, The Platonic representation hypothesis~\cite{platonic_hypothesis} posits that internal representation spaces scale toward a shared statistical model of reality, independent of architecture, objective, or modality. 
Recent empirical work has shown that dataset overlap and task similarity serve as the primary drivers of this alignment \cite{Li2025May}. However, this convergence may not be uniform; closely aligning with our own findings, \citet{revisiting_platonic} demonstrated that vision and language models frequently converge locally while maintaining global topological differences.

\paragraph{Comparing transformers and SSMs}
A growing body of work examines whether non-attention architectures, such as selective State-Space Models~(SSMs) like Mamba, converge on the internal representations used by transformers. Recent findings strongly support this universality hypothesis across multiple granularities. At the representation level, \citet{towards_universality} applied Sparse Autoencoders (SAEs) to demonstrate that both architectures extract a highly similar set of interpretable features. This convergence extends to specific semantic sub-domains; for instance, \citet{convergent_evo_numbers}

observed that both Mamba and transformers independently develop matching geometries for numerical tokens, achieving spectral and geometric convergence on periodic features. Furthermore, this alignment persists when analyzing operational mechanics: \citet{Yoo2026Apr} decomposed layer-wise updates across both model families and found that full updates are geometrically dominated by a highly aligned tokenwise transformation.

\paragraph{The geometry and anisotropy of latent spaces}
The geometric properties of these representations often dictate model utility and performance. In transformers, the self-attention mechanism is a known source of \textit{anisotropy}---a phenomenon where representations occupy a narrow cone in the latent space \cite{Gao2019Jul, Godey2024Jan, all_bark_no_bite}. While some models, such as the Pythia family, exhibit lower anisotropy than other transformers ~\cite{machina-mercer-2024-anisotropy}, we find they remain significantly more anisotropic than \ssms. This geometric structure is not merely a byproduct but a predictor of performance, as measures of isotropy correlate with LLM efficacy \cite{predictive_dispersion}. There has also been previous work which tried to learn about the subspaces responsible for specific attributes; \citet{low_dim_probe} demonstrated that many attributes reside in low-dimensional linear spaces in both BERT and ELMO.

\paragraph{Dynamic and layer-wise evolution}
Representational geometry is not static; it evolves significantly during training \cite{tracing_representations} and across layers~\cite{Wolfram2025Apr, 10.5555/3780338.3782559, 2nn_paper}. Previous work has also looked into similarities to tokens across contexts~\cite{bert_contextual} and across architectures~\cite{similarity_contextual_belinkov}.

\section{Conclusion}
In this work, we investigated the internal representations of \ssm and transformer architectures, analyzing their latent space utilization, geometric properties, and functional capacity. We found that \ssm representations are highly isotropic, whereas transformers are heavily anisotropic. In hybrid models, we observed architectural oscillations, where the representation space fluctuates between isotropic and anisotropic states across layers.

Despite these stark global geometric differences, both architectures exhibit surprising similarities. We demonstrated that their effective dimensionality—measured via autoencoders and rank-constrained bottlenecks—is remarkably similar, and the transformer's singular dominant direction does not inherently capture more conceptual information than \ssms' distributed space.

Structural alignment between architectures proved highly uneven across scope: high when restricted to individual semantic relations, but low over the general corpus. This local convergence, like the local convergence in CKA and \mknn more broadly, also emerges early in training, ahead of the slower-stabilizing global geometry.

Taken together, our results  indicate that transformers and \ssms are geometrically distinct and differ in how they organize latent space, yet converge on highly aligned local representational structures.

\section*{Limitations}
While our multi-scale analysis provides systematic insights into the representational mechanics of state-space, hybrid, and transformer architectures, we acknowledge the following boundaries to our scope:
\paragraph{Correlational vs.~causal analysis} 
Our evaluation of concept encoding relies primarily on rank-constrained probing, principal component ablations, and structural alignment metrics (CKA, \mknn). While these methods demonstrate that concept information can be decoded with equivalent linear capacity and share local manifold geometries, they do not establish a direct causal link to downstream generation \cite{probing_not_causal1, probing_not_causal2}.

\section{Acknowledgments}
This work was supported in part by the Israel Science Foundation (grant no. 2045/21) and by a research gift from Google.

This manuscript was drafted and refined with the assistance of large language models (LLMs). All content was reviewed and verified by the authors.

\bibliography{custom}

\appendix

\section{Hyperparameters for Experiments}
To ensure reproducibility, all code is publicly available and included as part of this ARR submission.

\begin{itemize}
    \item \textbf{Global Geometric Metrics (Section 3):} Evaluated on 1,500 samples from FineWeb-Edu with a maximum sequence length of 512, yielding a total of approximately 500K tokens.
    \item \textbf{Autoencoder Training (Section 4.1):} Trained for a single epoch on 10,000 FineWeb-Edu samples (maximum sequence length of 512) using a learning rate of 1e-4 and ReLU activations. It was evaluated over 1000 samples (maximum sequence length of 512).
    \item \textbf{Ablated Probing (Section 4.2):} Trained for 5 epochs with a learning rate of 1e-3 and an 80\%/20\% train-evaluation split. The training data was capped at a maximum of 10,000 samples per LAMA TREx relation.
    \item \textbf{ZCA-CKA (Section 5.1):} For FineWeb-Edu, metrics were computed over $\approx$10,000 tokens (derived from 100 samples of length 100). For LAMA TREx, we utilized the full dataset but sampled exactly one context sequence per unique factual triplet (subject, relation, object) to control for redundant semantic structures.
    \item \textbf{\mknn (Section 5.2):} Computed using 10,000 representations extracted from FineWeb-Edu.
\end{itemize}

\section{Additional Geometric Metrics}

\subsection{Global Geometric Metrics for Pythia Checkpoints}\label{sec:pythia_checkpoints}
\begin{figure*}[h]
    \centering
    \includegraphics[width=1\textwidth]{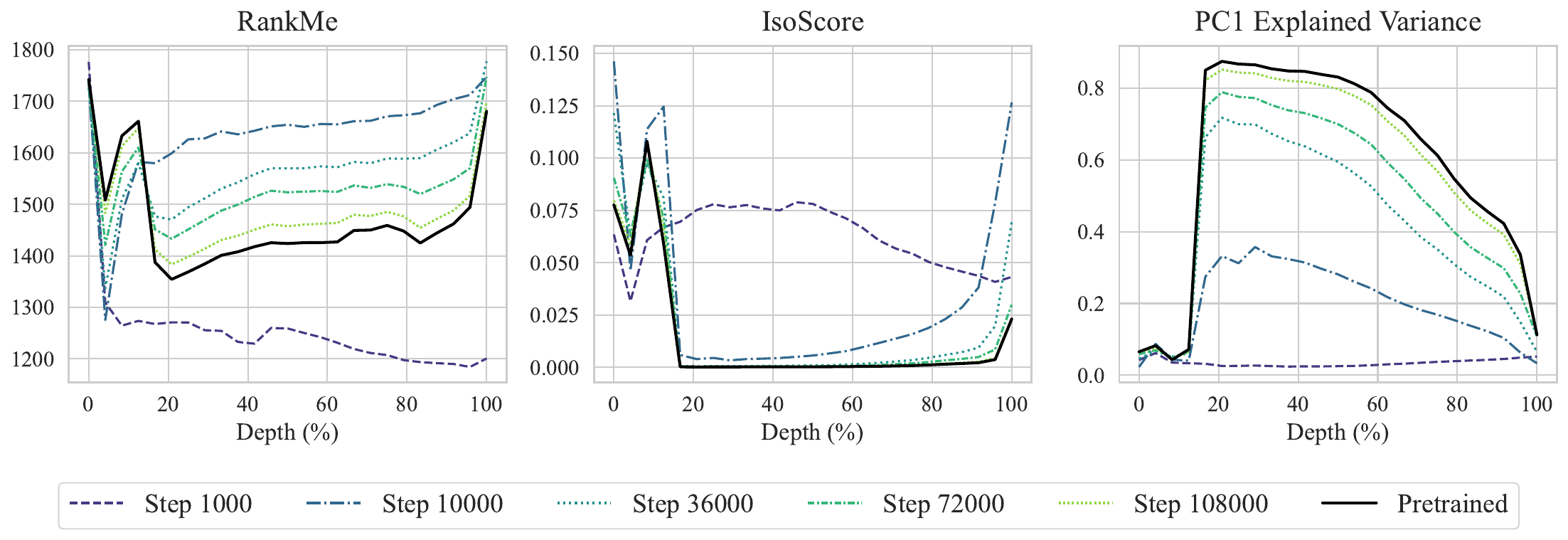}
    \caption{Layer-wise Geometric Analysis of Pythia-1.4B Checkpoints over $\approx$500K representations from FineWeb-Edu.}
    \label{fig:pythia_checkpoints}
\end{figure*}

We analyze Pythia's intermediate training checkpoints using the same metrics as described in \cref{sec:global_geom}. As shown in \cref{fig:pythia_checkpoints}, we find distinct optimization trajectories over the course of training. Specifically, RankMe exhibits a non-monotonic pattern, characterized by an initial increase followed by a subsequent decline. In contrast, IsoScore and PC1 explained variance converge monotonically toward their final values, with the primary stabilization occurring rapidly within the first 25\% of training steps.

\subsection{Global Geometric Metrics of Additional Models}\label{sec:appendix metrics}

\begin{figure*}[h]
    \centering
    \includegraphics[width=1\textwidth]{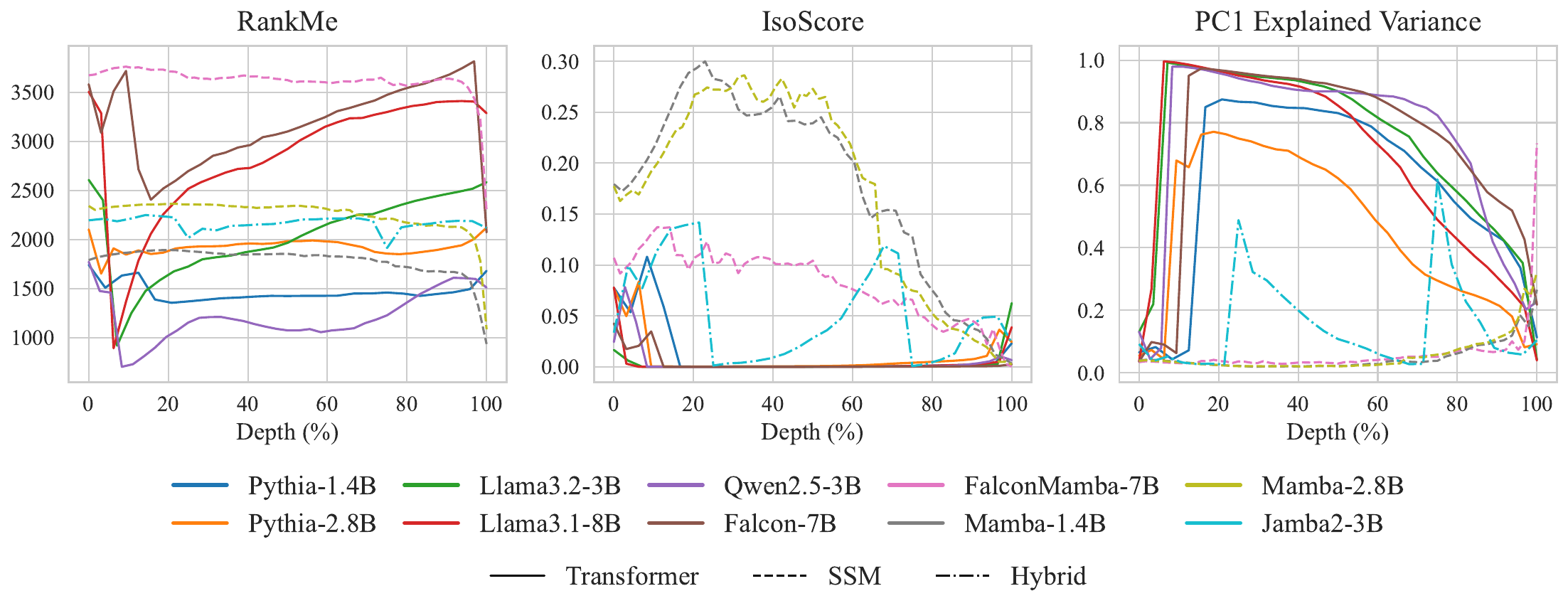}
    \caption{Layer-wise geometric analysis of state-space and transformer models over $\approx$500K representations from FineWeb-Edu. Transformers retain a single dominant axis across sizes and families, while Mamba's representation distribution remains uniform across different scales.}
        \label{fig:appendix_metrics}
\end{figure*}

We investigate the geometric metrics of additional models, namely for Llama \cite{llama_3} and Qwen \cite{qwen_2}. As shown in \cref{fig:appendix_metrics}, the empirical results indicate that while model scale inherently influences RankMe, \ssms consistently maintain a higher rank than Transformers when controlling for scale. Conversely, both IsoScore and PC1 explained variance remain largely invariant to scale.

\subsection{Variance Analysis of Geometric Metrics}
To verify the stability of our geometric metrics, we re-evaluate both models on 5 random seeds, each sampling a different subset of the FineWeb-Edu \texttt{sample-10BT} split, with each subset comprising $\approx 500K$ tokens. For each metric and layer, CV\% is computed across seeds; \cref{tab:geom_metrics_std} reports the mean of these values across layers, alongside the maximum layer-wise CV\% in brackets.

\begin{table}
\centering
\begin{tabular}{lcc}
\toprule
\textbf{Metric} & \textbf{Pythia-1.4B} & \textbf{Mamba-1.4B} \\
\midrule
RankMe & $0.13\% \, [0.18\%]$ & $0.11\% \, [0.34\%]$ \\
IsoScore & $0.52\% \, [0.98\%]$ & $0.50\% \, [1.\%]$ \\
PC1 Exp. Var. & $0.18\% \, [0.48\%]$ & $0.36\% \, [0.77\%]$ \\
\bottomrule
\end{tabular}
\caption{Stability of representation geometry metrics across 5 random seeds, measured via coefficient of variation (CV\%). For each metric, we report the mean CV\% across layers, with the maximum layer-wise CV\% in brackets. Lower values indicate greater stability.}
\label{tab:geom_metrics_std}
\end{table}

\section{Samples from LAMA TREx}\label{sec: lama_trex_samples}
\cref{tab:lama_trex_samples} presents samples from the LAMA TREx dataset \cite{lama_dataset}, which spans 41 distinct Wikidata relation types. The dataset pairs Wikidata triples (subject, relation, object) with aligned natural language sentences from Wikipedia.

\begin{table*}[ht]
\centering
\small
\renewcommand{\arraystretch}{1.3}
\begin{tabularx}{\textwidth}{l l X}
\toprule
\textbf{Relation} & \textbf{ID} & \textbf{Sentence} \\ 
\midrule
P136 (genre) & 19918 & The Simpsons is the longest-running American \underline{sitcom}, the longest-running American animated program, and in 2009 it surpassed Gunsmoke as the longest-running American scripted primetime television series. \\

P30 (continent) & 336 & Historical records of Western culture in \underline{Europe} begin with Ancient Greece and Ancient Rome. \\

P101 (field of work
) & 5656 & Brouwer's fixed-point theorem is a fixed-point theorem in \underline{topology}, named after Luitzen Brouwer. \\

P106 (occupation) & 15826 & That \underline{astronaut}, Dumitru Prunariu is today's president of Romanian Space Agency.\\

P37 (official language) & 16094 & Australian English (AusE, AuE, AusEng, en-AU) is a major variety of the \underline{English language} and is used throughout Australia. \\

P31  (instance of) & 2448 & Another version of Hahn–Banach theorem is known as Hahn–Banach separation \underline{theorem} or the separating hyperplane theorem, and has numerous uses in convex geometry.\\

\bottomrule
\end{tabularx}
\caption{Examples of LAMA TREx instances across distinct relation categories. Underlined words indicate the object.}
\label{tab:lama_trex_samples}
\end{table*}

\section{Additional Results for Model Capacity Limits}\label{subsec: appendix_ae_full}
\begin{figure}[h!]
    \centering
    \includegraphics[width=\linewidth]{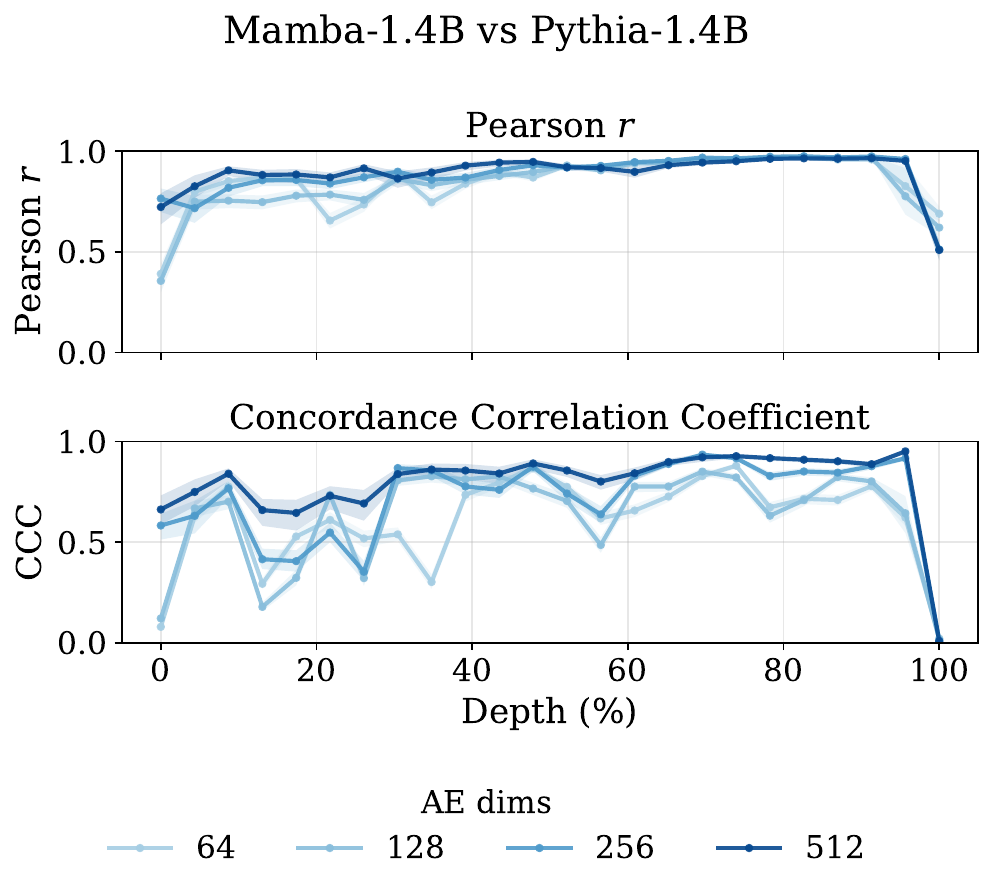}
    \caption{Pearson and CCC correlation between the KL-divergences of Mamba-1.4B and Pythia-1.4B.}
    
    \label{fig:ae_correlation}
\end{figure}

\begin{figure*}[h]
    \centering
    \includegraphics[width=1\textwidth]{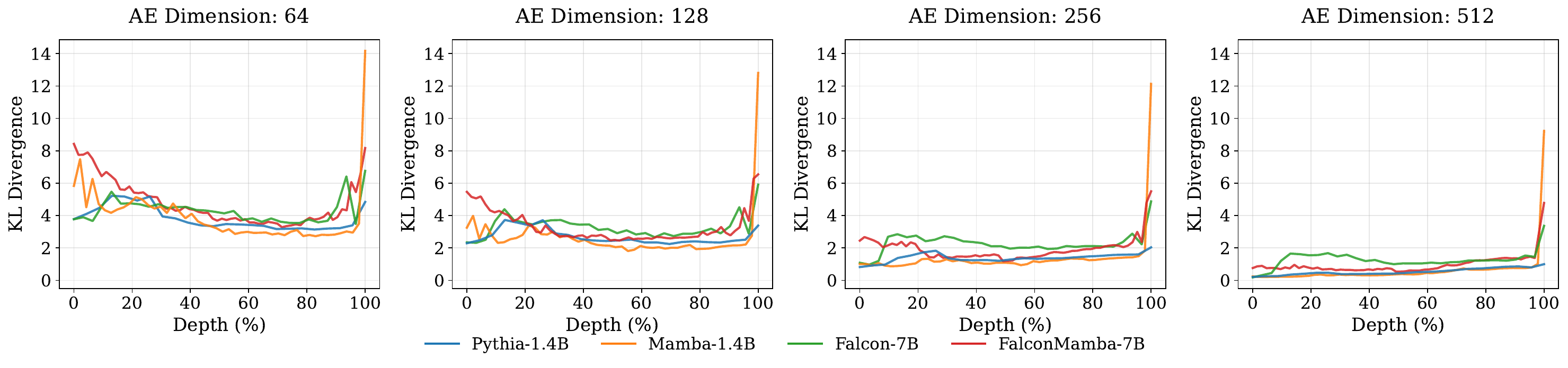}
    \caption{Per-layer KL-divergence between original and reconstructed representations across additional constraint dimensions.}
    \label{fig:ae_kl_full}
\end{figure*}

\begin{figure*}[h]
    \centering
    \includegraphics[width=1\textwidth]{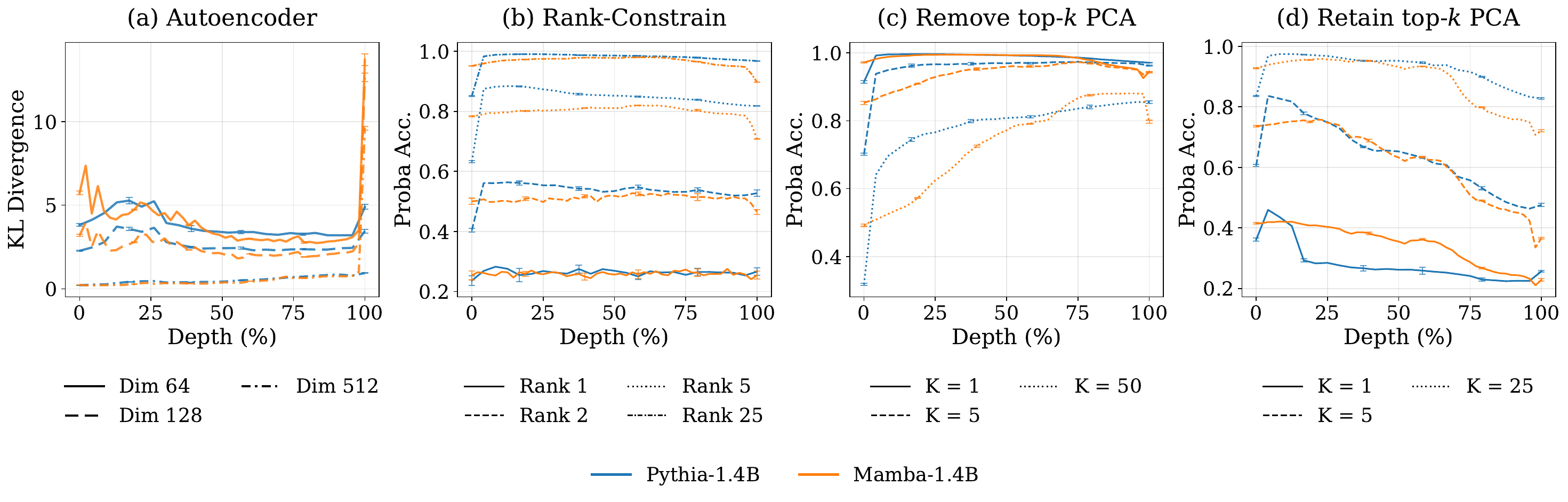}
    \caption{Per-layer KL-divergence between original and reconstructed, with mean $\pm$ standard deviation over 5 random seeds.}
    \label{fig:ae_kl_seed}
\end{figure*}

We extend the experiment in~\cref{subsec:autoencoder} to additional autoencoder rank constraints $r$. As shown in~\cref{fig:ae_kl_full} (additional rank constraints) and~\cref{fig:ae_kl_seed} (mean $\pm$ std. over 5 seeds), the degradation pattern remains consistent across models for every constraint tested. To quantify this agreement, we compute the Pearson correlation and the Concordance Correlation Coefficient (CCC) between the per-layer KL-divergence curves of the two architectures (\cref{fig:ae_correlation}, \cref{tab:ae_correlation}). Correlation is consistently high (Pearson $>0.84$ across all $r$) and broadly increases with rank, from $0.857$ at $r=64$ to $0.911$ at $r=512$; CCC follows the same trend, rising from $0.638$ to $0.830$. This indicates that agreement between architectures strengthens as the bottleneck is relaxed, i.e., degradation curves are most alike where capacity constraints are less severe.

\begin{table}[h!]
\centering
\begin{tabular}{lcc}
\toprule
AE Dimension & Pearson & CCC \\
\midrule
64  & 0.857 $\pm$ 0.027 & 0.638 $\pm$ 0.041 \\
128 & 0.849 $\pm$ 0.028 & 0.658 $\pm$ 0.045 \\
256 & 0.901 $\pm$ 0.015 & 0.744 $\pm$ 0.037 \\
512 & 0.911 $\pm$ 0.012 & 0.830 $\pm$ 0.020 \\
\bottomrule
\end{tabular}

\caption{Mean Pearson and Lin's CCC between per-layer KL-degradation curves of Mamba-1.4B and Pythia-1.4B, per rank constraint $r$ (averaged across layers). The final SSM layer is excluded: its KL divergence is disproportionately high, unlike the corresponding Transformer layer (\cref{subsec:autoencoder}).}
\label{tab:ae_correlation}
\end{table}

\section{Additional CKA Results}

\begin{figure}[h!]
    \centering
    \includegraphics[width=\linewidth]{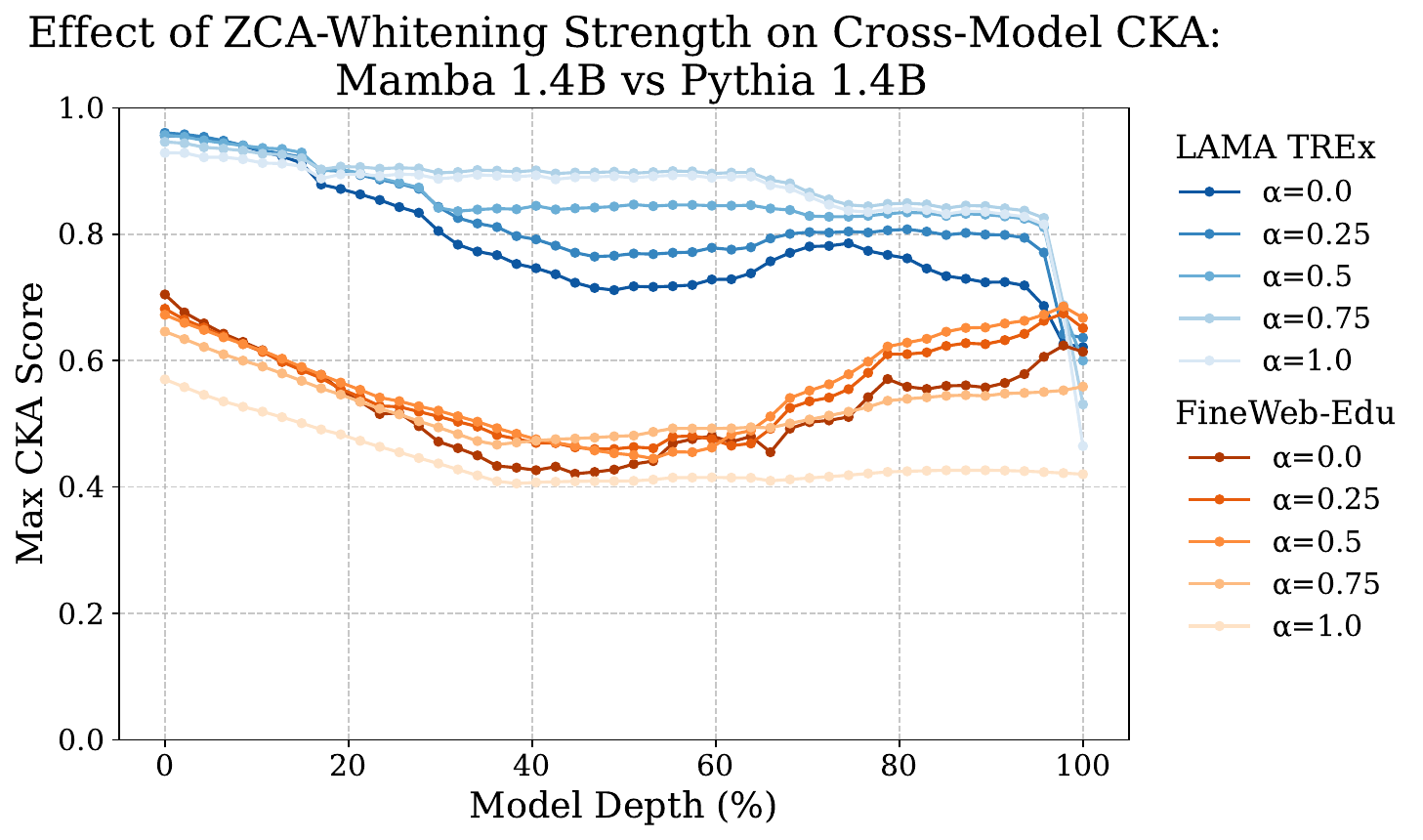}
    \caption{Maximum pairwise CKA similarity across all Mamba-1.4B, Pythia-1.4B layer pairs, plotted as a function of ZCA-whitening strength $\alpha$, for LAMA TREx (mean over relations) and FineWeb-Edu.}
    \label{fig:ZCA_alpha}
\end{figure}

In \cref{sec:local_similarity}, ZCA-whitening was used to address the fact that the two models have markedly different spectral properties. To test whether the local/global alignment gap reported there is a genuine property of the representations rather than an artifact of this spectral mismatch, we recompute the maximally-similar-layer CKA under ZCA-whitening at several strengths $\alpha$ by computing $\Sigma^{-\alpha/2}$ over both LAMA TREx (mean across relations) and FineWeb-Edu (\cref{fig:ZCA_alpha}).
Even without whitening ($\alpha=0$), LAMA TREx representations are substantially more aligned than FineWeb-Edu representations (layer mean $0.78$ vs. $0.52$), indicating that the gap is not an artifact of the ZCA whitening.

To confirm the adequacy of a 10K sample size for ZCA-CKA, we computed the coefficient of variation (CV\%) for Mamba-1.4B versus Pythia-1.4B across varying seeds. On FineWeb-Edu, using 2.5K-sample subsets, we observed a low CV\% across all layer pairs (mean = 0.15\%, max = 1.34\%). We found similarly low variance over 5 seeds from the LAMA TREx split (mean = 0.56\%, max = 3.31\%). These results indicate that our chosen sample size is highly stable.

\section{Additional Results for \mknn}\label{sec:mknn_falcon}
We further validate the robustness of our \mknn results from \cref{subsec:mknn} along two axes: generalization to unseen architectures, and stability with respect to sampling.

\begin{figure}[h]
    \centering
    \includegraphics[width=\linewidth]{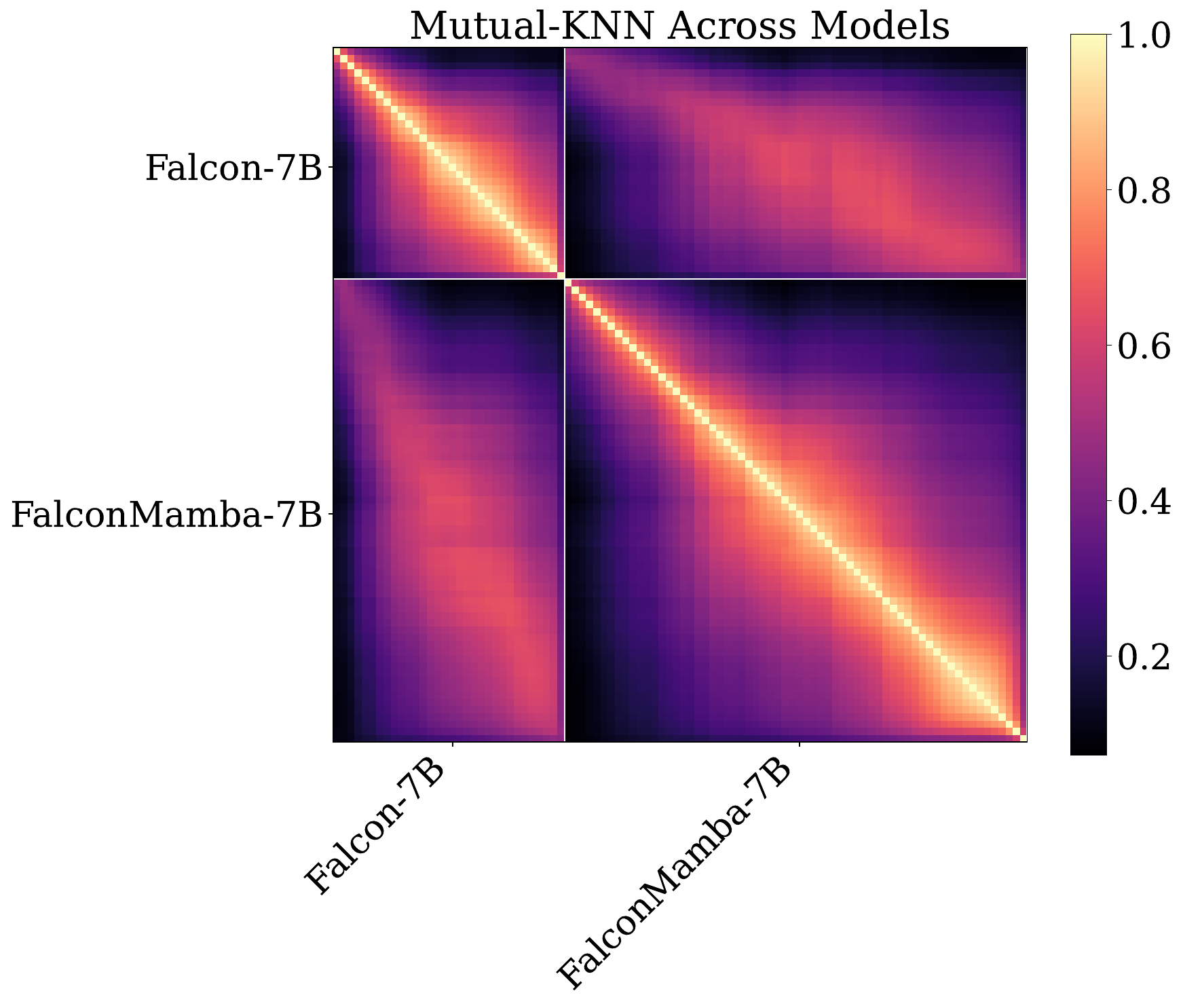}
    \caption{\mknn across models and layers for Falcon-7B and FalconMamba-7B. We find that the local manifold of representations is similar across models at similar depths.}
    \label{fig:mknn_falcon}
\end{figure}

\paragraph{Generalization to new architectures.}
We repeat the analysis on Falcon-7B and FalconMamba-7B, an architecture pair not included in the original comparison. As shown in \cref{fig:mknn_falcon}, we observe the same trend as before: local neighborhood topology remains highly aligned at corresponding relative depths, confirming that this alignment is not specific to the Mamba/Pythia family.

\paragraph{Stability with respect to sampling.}
To verify that our sample size is sufficient for stable estimates, we recompute \mknn on five independently drawn batches of 2,000 samples each, for the Mamba and Pythia models (1.4B and 2.8B). For each of the layer-pair comparisons, we compute the mean and standard error across the five batches ($n=5$). The mean SEM across all pairs was $0.0047$ (range: $0.0003$--$0.0113$), and the mean standard deviation was $0.0104$ (range: $0.0008$--$0.0253$), confirming that our results are stable with respect to sampling variation.


\end{document}